\documentclass[11pt]{article}

\usepackage[preprint]{acl}

\usepackage{times}
\usepackage{latexsym}

\usepackage[T1]{fontenc}

\usepackage[utf8]{inputenc}

\usepackage{microtype}

\usepackage{inconsolata}

\usepackage{graphicx}

\usepackage{booktabs}
\usepackage{amssymb}
\usepackage{pifont}

\usepackage[normalem]{ulem}

\title{A Computational Approach to Language Contact -- A Case Study of Persian}

\author{
  \textbf{Ali Basirat\textsuperscript{1}}\thanks{All authors contributed equally to this work.},
  \textbf{Danial Namazifard\textsuperscript{2}},
  \textbf{Navid Baradaran Hemmati\textsuperscript{3}}
\\
\\
  \textsuperscript{1}Centre for Language Technology (CST), University of Copenhagen\\
  \textsuperscript{2}University of Tehran\\
  \textsuperscript{3}Certified Translation Agency No. 1141, Mashhad, Iran
\\
  \small{
    \textbf{Correspondence:} \href{mailto:alib@hum.ku.dk}{alib@hum.ku.dk}
  }
}

\begin{document}
\maketitle
\begin{abstract}
We investigate structural traces of language contact in the intermediate representations of a monolingual language model. Focusing on Persian (Farsi) as a historically contact-rich language, we probe the representations of a Persian-trained model when exposed to languages with varying degrees and types of contact with Persian. Our methodology quantifies the amount of linguistic information encoded in intermediate representations and assesses how this information is distributed across model components for different morphosyntactic features. The results show that universal syntactic information is largely insensitive to historical contact, whereas morphological features such as \textsc{Case} and \textsc{Gender} are strongly shaped by language-specific structure, suggesting that contact effects in monolingual language models are selective and structurally constrained.
\end{abstract}

\section{Introduction}
Language contact is a primary driver of language change and structural diversification \cite{vanCoetsem2000linguistic}. When speakers of different languages come into sustained interaction through, for example, trade, migration, conquest, or cultural exchange, their languages influence one another in ways that can range from lexical borrowing to deep grammatical restructuring \citep{thomason1988language,matras2009language}. 
This can appear as \emph{matter borrowing}, where the morphological material and its phonological shape are replicated, and/or as \emph{pattern borrowing}, where the organization, distribution, and mapping of grammatical or semantic meaning patterns are replicated \citep{Sakel+2007+15+30}. 


The study of language contact requires a holistic perspective on linguistic structure that spans multiple levels of representation \cite{matras2009language}, a perspective that is difficult to obtain solely through traditional analytical methods. Large language models (LLMs) have opened new avenues for investigating linguistic phenomena by enabling the study of their intermediate representations. Although LLMs are not designed to model diachrony, their internal representations often reflect patterns of cross-linguistic similarity and genealogical relatedness that emerge implicitly from large-scale training data \cite{veeman-etal-2020-cross,chi-etal-2020-finding}. 

This study investigates how language change might have been reflected in the intermediate representation of a monolingual Persian language model. The case of Persian (Farsi) provides a particularly compelling testbed. Although genealogically Indo-European, Persian has undergone extensive contact with languages from multiple families, most notably Turkic languages, as well as Arabic, English, French, Russian, and Hindi \citep{windfuhr2009iranian}. These contact dynamics have influenced Persian at several linguistic levels. However, it remains unclear whether such changes are detectable, quantifiable, and linguistically interpretable within the internal representations of a monolingual language model, such as ParsBERT \citep{farahani2021parsbert}. 

We study language change at the morpho-syntactic level based on linguistic annotations from Parallel UD treebanks \citep{zeman-etal-2017-conll} for a group of languages at various degrees and types of language contact. We adopt an information-theoretic approach to measure the amount of usable information \citep{Xu2020A} that intermediate representations of ParsBERT encode about the linguistic features of the target languages. We further perform an attribution analysis \citep{tang-etal-2024-language} to identify those components of the representations that contribute most strongly to this encoding.

Our findings contribute to a growing body of research on typology-aware NLP \citep{bender-2009-linguistically,ponti-etal-2019-modeling}, offering evidence that monolingual language models encode selective and structurally constrained traces of language contact. Specifically, we show that contact effects are more readily reflected in morphological and constructional patterns aligned with the training language, while universal syntactic categories remain largely robust to contact-induced variation. By combining information-theoretic probing with attribution analysis, this work provides a principled computational framework for investigating classical questions in contact linguistics and sheds light on the limits and possibilities of using neural language models as tools for studying language change.
\section{Related Work}
\label{sec:related-work}

Language contact has been extensively studied in functional, historical, and typological linguistics, with a particular focus on how sustained interaction between speech communities leads to lexical, morphological, and syntactic change \citep{thomason1988language,matras2009language}. Persian constitutes a well-documented case of long-term language contact: centuries of interaction with Semitic, Turkic, Indic, and European languages have left pervasive traces across multiple linguistic levels \citep{windfuhr2009iranian,JohansonCsato2021}. While these contact-induced effects have been described in detail from diachronic and descriptive perspectives, they have not yet been examined using computational models of language. In particular, it remains unclear how language contact phenomena are reflected in the internal representations of neural language models trained on Persian. To the best of our knowledge, this work presents the first computational study of Persian language contact at the level of neural representations.

Our work is also related to the growing literature on probing neural language models for linguistic structure. Prior research has investigated how morphology \citep{stanczak-etal-2022-neurons}, syntactic structure \citep{tenney-etal-2019-bert,hewitt-manning-2019-structural}, and cross-linguistic typological properties \citep{ponti-etal-2019-modeling} are encoded in contextualized representations. Grammatical gender, in particular, has been widely used as a test case for cross-linguistic generalization in neural models \citep{veeman-etal-2020-cross,schroter-basirat-2025-universal}. We extend this line of work by probing the intermediate representations of a monolingual Persian language model, with a focus on how linguistic features associated with contact languages are encoded. By doing so, we connect descriptive insights from language contact research with contemporary computational approaches to representation learning.

\section{Methodology}

Our goal is to determine whether a monolingual Persian model encodes latent structural traces that reflect Persian's long-standing contact with other languages. To this end, we examine whether representations learned exclusively from Persian data exhibit systematic affinities to linguistic characteristics of contact languages.

We pass a set of sentences sampled from a contact language into a Persian language model. For each token, we extract the intermediate representations produced by each of the embedding and Transformer layers of the model. We, then, adopt two techniques to analyze these intermediate representations (embeddings): (i) an information-theoretic probe that quantifies how much linguistic information is encoded in the representations, and (ii) a complementary attribution analysis that identifies which components of the representations contribute most strongly to this encoding.

The probing framework is based on \emph{variational usable information} $I_{\mathcal{V}}(X \rightarrow Y)$ \citep{Xu2020A}, which estimates the amount of information that a random variable $X$ contains about a target random variable $Y$. In our study, $X$ corresponds to embedding vectors extracted from a given intermediate layer of the language model when processing a token in context, and $Y$ represents a linguistic property of the token (e.g., language identity, UPOS tag, \textsc{Case}, or \textsc{Gender}). Following \citet{basirat-2025-multilingual}, we normalize $I_{\mathcal{V}}(X \rightarrow Y)$ by marginal entropy $H(Y)$ of the target feature. This normalization yields a dimensionless metric $\hat{I}_\mathcal{V} \in [0,1]$ that reflects the proportion of the total uncertainty in $Y$ that can be recovered from the representation. 

High usable-information for a linguistic property in a test language indicate that ParsBERT encodes systematic cues relevant to the realization of that property, suggesting structural alignment between the test language and Persian for the corresponding feature. Conversely, low usable-information scores reflect limited overlap between the feature’s realization in the test language and Persian.

The attribution analysis quantifies how linguistic information about a target variable $Y$ is distributed across elements of $X$ using \emph{Language Activation Probability Entropy} (LAPE) \citep{tang-etal-2024-language}. LAPE assigns each element of $X$ a score based on the entropy of its activation distribution across different \emph{conditions}, corresponding to the linguistic properties of the processed tokens. Low LAPE scores indicate high selectivity, meaning that an element responds preferentially to a specific condition, whereas high scores reflect more uniform or non-discriminative activation patterns. To identify \emph{condition-selective} elements, we rank all elements by their LAPE scores and retain only those within the lowest percentile, discarding elements that are rarely active. Each retained element is then assigned to the condition for which it exhibits the strongest activation preference.

We summarize LAPE by reporting the \emph{total number of elements} across all layers assigned to each condition.\footnote{We also report layer-wise LAPE scores in Appendix~\ref{appx:lape}.} The counts primarily reflect \emph{localization} rather than the overall presence or absence of information for a condition. In our interpretation, larger counts suggest that selectivity for a condition is distributed across more elements, whereas smaller counts suggest that selectivity is concentrated in fewer units. Importantly, small counts do not imply that the model cannot encode a condition; the signal may instead be distributed across shared elements in a way that is not captured by a small set of highly selective units. 

\section{Experiment Setup}

\subsection{Model}
Our experiments are conducted on ParseBERT \citep{farahani2021parsbert}, a monolingual encoder-only model \citep{devlin2019bert} consisting of 12 transformer encoder layers, each with a hidden size of 768 (i.e., the dimensionality of $X$). The tokenizer is a WordPiece model trained on Persian-script text. The model is pretrained on approximately 3.9 billion tokens of Persian text compiled from multiple large-scale sources, spanning genres such as encyclopedic, journalistic, conversational, and technical, and written in both standard and contemporary Persian writing systems. This breadth is particularly relevant for studying language-contact phenomena in the model, as many contact-induced lexical and syntactic patterns (e.g., English and French loanwords, Arabic learned vocabulary, Turkic colloquialisms) appear prominently in modern Persian online text.

Because ParsBERT's training is strictly monolingual, any alignment between its internal representations and the structural patterns of contact languages cannot result from direct exposure to those languages during pretraining. Instead, such patterns must arise from:
\begin{itemize}
    \item structural features of Persian itself that reflect historical contact;
    \item statistical imprint of borrowings, calques, and hybrid constructions;
    \item universal linguistic properties shared across languages.
\end{itemize}
These factors motivate our choice of linguistic characteristics analyzed in the remainder of this paper.

\subsection{Data: Parallel Universal Dependencies}
We employ the Parallel Universal Dependencies (PUD) treebanks \citep{zeman-etal-2017-conll,nivre-etal-2016-universal} as our cross-linguistic evaluation data. PUD consists of sentence-level translations of the same source texts of 1000 news and Wikipedia sentences, each annotated according to the Universal Dependencies (UD) framework. 

The consistent annotation of the data enables a cross-lingual comparison of the results, and the alignment between the sentences minimizes domain-induced variation in the evaluation. Consequently, differences in the analyses across languages more reliably reflect underlying linguistic structure--morphology and syntax--rather than divergences in domain or stylistic conventions.

\subsection{Test Languages}
Out of the 21 languages available in PUD, we focus on 8 languages with different types and degrees of contact with Persian (i.e., historical, modern, regional, or minimal). To disentangle contact-driven effects from general typological similarity, we additionally include one language with moderate contact and one with minimal or no contact as comparative controls. Table~\ref{tab:data_properties} outlines the languages and statistics of the datasets used in our experiments. 
Below, we outline the degree and nature of contact between Persian and selected languages.
\begin{table*}[t]
\centering
\resizebox{\textwidth}{!}{
\begin{tabular}{llllllll}
\toprule
\textbf{Lang (ISO)} & \textbf{Family} & \textbf{Morph.} & \textbf{\#Tok} & \textbf{Writing System} & \textbf{Case} & \textbf{Gender} & \textbf{Contact Type} \\
\midrule
Arabic (ar)   & AA: Semitic     & Templatic        & 21K & Consonantal         & Rich (3)          & M/F          & Historical (Major) \\
English (en)  & IE: Germanic    & Analytic         & 21K & Alphabetic          & Pronominal        & Pronominal   & Modern (Global) \\
French (fr)   & IE: Romance     & Fusional         & 24K & Alphabetic          & Pronominal        & M/F          & Modern (19–20 c.) \\
German (de)   & IE: Germanic    & Fusional         & 21K & Alphabetic          & Rich (4)          & M/F/N        & Minimal \\
Hindi (hi)    & IE: Indo-Aryan  & Fus./Analyt.     & 23K & Syllabic     & Postpositional    & M/F          & Historical (Prestige) \\
Japanese (ja) & Japonic         & Agglutinative    & 28K & Logographic+Syllabic     & Particles         & \ding{55}    & Minimal \\
Russian (ru)  & IE: Slavic      & Fusional         & 19K & Alphabetic       & Rich (6)          & M/F/N        & Regional (Modern/20 c.) \\
Turkish (tr)  & Turkic          & Agglutinative    & 17K & Alphabetic          & Rich (6+)         & \ding{55}    & Historical (Areal) \\
\midrule
Persian (fa)  & IE: Iranian     & Analytic/LVC     & --- & Consonantal   & \ding{55}         & \ding{55}    & --- \\
\bottomrule
\end{tabular}
}
\caption{Linguistic properties of the test languages. Abbreviations: IE = Indo-European, AA = Afro-Asiatic, Morph.\ Type = morphological type, LVC = light-verb construction. Case systems are summarized as: Rich ($n$) = languages with $n$ core grammatical cases; Postpos = postpositional case marking; Particles = case marking via grammatical particles; (L)~Pron = case distinctions limited to pronouns. Gender systems are marked as M/F/N = masculine/feminine/neuter or \ding{55} for no grammatical gender. Contact Type indicates the historical, modern, regional, or minimal degree of contact each language has had with Persian.}
\label{tab:data_properties}
\end{table*}

\paragraph{Turkish (tr).}
Persian and Turkic languages have been in continuous contact for nearly a millennium, beginning in the 11th century \citep{JohansonCsato2021}. 
The interaction was bidirectional: Ottoman Turkish absorbed extensive Persian vocabulary and literary forms \citep{lewis1999turkish}, while Persian borrowed pastoral, military, and administrative terms from Turkic languages \citep{Johanson2006Structural}. 
Among the PUD languages, Turkish represents the most intense, long-lasting, and structurally consequential contact relationship with Persian. Turkish adopted the Latin alphabet in the late 1920s.

\paragraph{Arabic (ar).} 
Following the Arab conquest of Iran in the 7th century CE, Persian underwent extensive borrowing from Arabic, especially in religion, administration, scholarship, and abstract vocabulary \citep{windfuhr2009iranian}. 
Although Persian and Arabic remain typologically distinct, Arabic is one of the most prominent sources of borrowed vocabulary in Persian, especially in formal and literary domains. Arabic uses the Abjad writing system. 

\paragraph{English (en).} 
English has exerted a strong influence on Persian primarily since the 20th century, driven by globalization, science, technology, education, and media circulation \citep{hosseini2020english}. Most influence is lexical, with widespread borrowing or calquing in technical domains. English has also assimilated Persian-origin words through multiple intermediaries, including \emph{bazaar}, \emph{caravan}, \emph{pajamas}, and \emph{checkmate} \citep{campbell2013historical}. English uses the Latin alphabet.

\paragraph{French (fr).} 
French was a major prestige language in Iran during the 19th and early 20th centuries, influencing administrative, legal, educational, and military terminology \citep{amanat2017iran}. Borrowings remain visible in modern Persian: \emph{šofor} (chauffeur), \emph{telviziōn} (télévision), \emph{mersi} (merci), among others. Although its influence has diminished today, French played a central role in Iran's modernization and lexical expansion. French uses the Latin alphabet.

\paragraph{Russian (ru).} 
Russian influence on Persian is strongest in northern dialect regions and in Tajik, which was deeply shaped by Russian during the Soviet period \citep{windfuhr2009iranian}. Borrowings appear in domains such as politics, science, technology, and daily life. Russian also affected Persian-speaking communities in the Caucasus (Azerbaijan, Dagestan) through bilingualism and administrative contact. While substantial, Russian contact is more regionally and temporally bounded compared with Turkish and Arabic, making it an appropriate \emph{moderate-contact} control. Russian uses the Cyrillic alphabet.

\paragraph{Hindi (hi).} 
Persian served as the administrative, literary, and court language across much of northern India for nearly seven centuries (ca.\ 1200--1857) \citep{alam2004persianate}. As a result, modern Hindi and especially Urdu contain thousands of Persian loanwords and exhibit syntactic and phraseological calques \citep{comrie2018major}. 

However, similarities between Persian and Hindi are not solely the outcome of prolonged contact. 
The two languages descend from the same subbranch of Indo-European, and thus share inherited lexical, morphological, and syntactic features traceable to earlier stages of Indo-Iranian. 
Nevertheless, in the historical period, the direction of influence is primarily Persian $\rightarrow$ Indo-Aryan, rather than the reverse. Hindi uses the Devanagari writing system.

\paragraph{Japanese (ja).} 
Japanese has no direct historical or areal contact with Persian. 
A small number of Persian-origin terms reached Japanese indirectly, typically via English or global trade (e.g., \emph{pajama}, \emph{caravan}) \citep{campbell2013historical}. 
The contact is therefore minimal, recent, and mediated, making Japanese a suitable \emph{no-contact baseline}. Japanese uses a mixed writing system combining logographic and syllabic scripts, known as Kanji and Kana, respectively.

\subsection{Linguistic Annotations}
We investigate language contact on (i) \emph{language-specific representations}, (ii) \emph{the encoding of universal syntax}, and (iii) \emph{morphological features absent from Persian}. Our analysis relies on token-level PUD annotations ($Y$) and the corresponding intermediate representations of ParsBERT ($X$). 

Language-specific representations are assessed by predicting each token's source language ($Y$) from its intermediate representation ($X$) at each layer of ParsBERT \citep{basirat-2025-multilingual}. The investigation of universal syntax is based on the Universal Part-of-Speech (UPOS) tags in the Universal Dependencies framework. Finally, the study of morphological contact is based on the features absent in Persian but attested in several contact languages. We restrict this analysis to \textsc{Case} and \textsc{Gender}, both of which are absent in Persian.\footnote{Persian is not a canonical case-marking language; however, the object marker \emph{-r\={a}} is sometimes analyzed as a differential or residual accusative case rather than a full case marker.} Language-specific information about these annotations is provided in Table~\ref{tab:data_properties}.

\section{Results}

This section reports the results of our probing and attribution analyses across all tasks. A detailed layer-wise analysis of the LAPE scores is provided in Appendix~\ref{appx:lape}.

\subsection{Language Identification}
\begin{figure}
    \centering    \includegraphics[width=\linewidth]{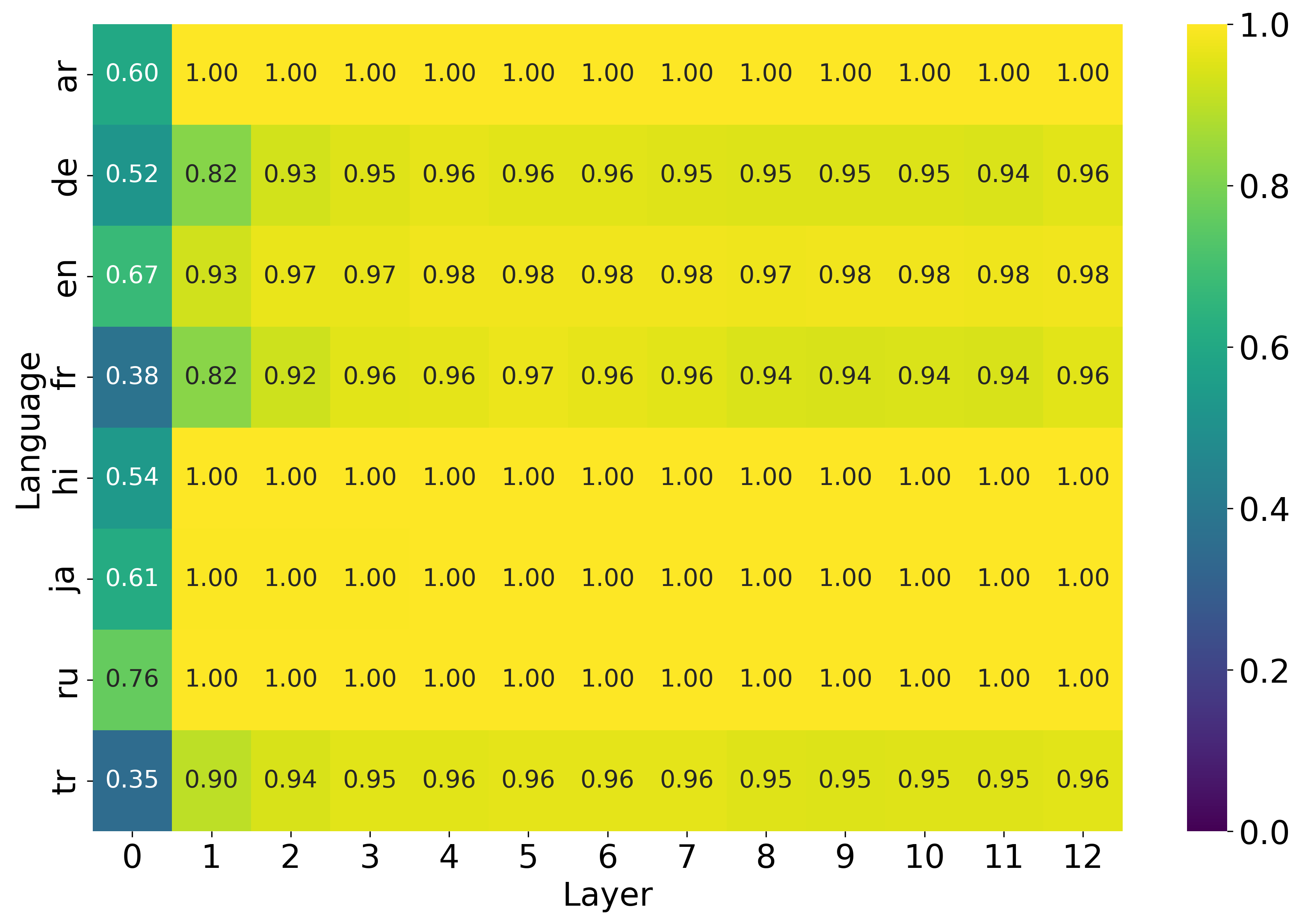}
    \caption{Normalized variational usable information ($\hat{I}_\mathcal{V}$) for language identification.}
    \label{fig:lang_id_nuvi}
\end{figure}

Figure~\ref{fig:lang_id_nuvi} presents a heatmap of usable information ($\hat{I}_\mathcal{V}$) for language identification across languages and layers. The results show that ParsBERT encodes substantial information about the identity of non-Persian languages, despite being trained exclusively on Persian data. The information gain increases across layers, peaking right after the third layer, and then stabilizes in higher layers.

The model is most informative about Arabic, Hindi, Japanese, and Russian. This likely reflects the significant, mainly lexical, influence of Arabic on Persian, the dual historical and genealogical relationship with Hindi, and the regional contact with Russian in northern Iran. The model's identification of Japanese remains unclear; further investigation is needed to clarify this.

A moderate amount of usable information is observed for English, and French, German, and Turkish exhibit lower values. The relatively low usable information for Turkish is particularly notable, given the extensive historical contact between Persian and Turkic languages. This may suggest that while Persian has absorbed substantial Turkic vocabulary, the deeper morpho-syntactic patterns of Turkish diverge more significantly from those encoded in ParsBERT's representations. 

\begin{figure}
    \centering
    \includegraphics[width=\linewidth]{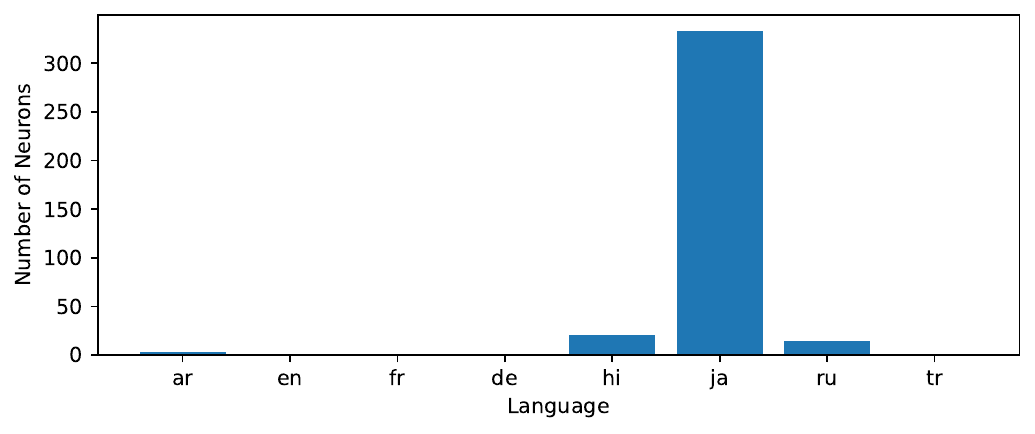}
    \caption{The LAPE scores for each language.}    \label{fig:lang_id_lape_total}
\end{figure}

Figure~\ref{fig:lang_id_lape_total} summarizes LAPE scores associated with each language. The distribution is highly skewed: no neurons are assigned to English, French, German, or Turkish under the LAPE criterion, while only a small number are associated with Arabic, Hindi, and Russian. In contrast, Japanese accounts for by far the largest score.

The results suggest that the language signal for Japanese is more readily isolated into many highly selective units, while for several other languages it is either weaker under LAPE’s selectivity criterion or remains expressed in shared (non-language-specific) features \citep{tang-etal-2024-language}. The prominence of Japanese may be related to its strong script/orthographic mismatch with Persian and the resulting low subword overlap under a Persian-trained WordPiece tokenizer, leading to more separable activation patterns associated with Japanese.

Overall, while the information analysis demonstrates that substantial language-identifying information is distributed across representations for several non-Persian languages, LAPE reveals that this information is rarely localized into compact, language-specific features. From a language-contact perspective, these findings suggest that historical contact and lexical influence contribute to distributed representational overlap, whereas strong orthographic and typological divergence--rather than contact per se--facilitates the emergence of localized language-specific features.

\subsection{UPOS Identification}

\begin{figure}
    \centering
    \includegraphics[width=\linewidth]{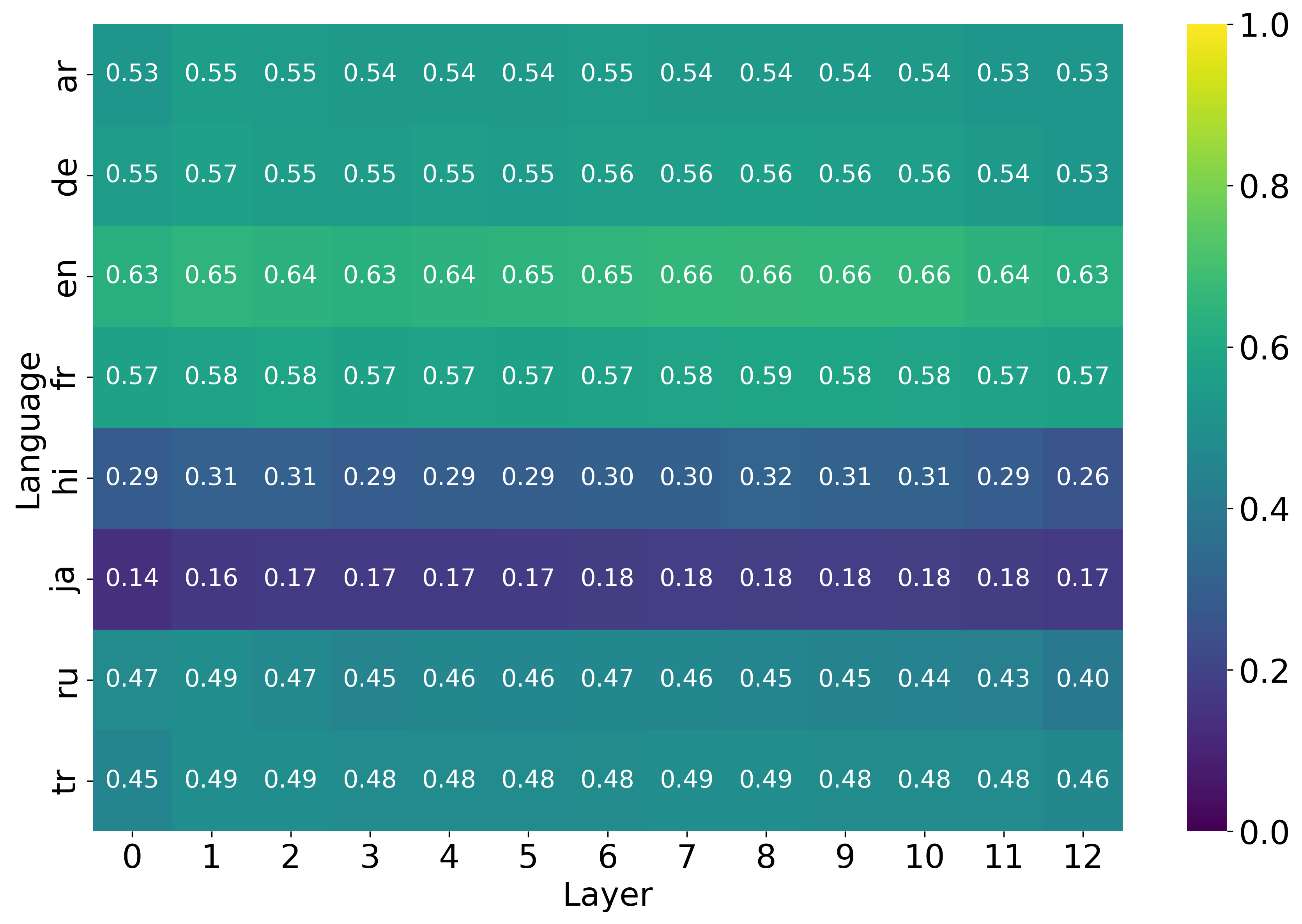}
    \caption{Normalized variational usable information ($\hat{I}_\mathcal{V}$) for UPOS identification.}
    \label{fig:upos_nuvi}
\end{figure}

Figure~\ref{fig:upos_nuvi} presents usable information values for UPOS identification across languages and layers. The information gain remains largely stable with respect to layer depth for all languages. English yields the highest score, followed by French, German, and Arabic, while Japanese and Hindi exhibit substantially lower values. Turkish and Russian fall in an intermediate range.

The results do not reveal a meaningful relationship between the degree of historical contact with Persian and the amount of recoverable UPOS information. Neither typological similarity to Persian (e.g., word order or morphological type) nor distributional similarity at the level of UPOS tag frequencies appears to be a dominant factor in explaining the observed variation in usable information. It supports the interpretation that universal syntactic representations for UPOS are largely insensitive to contact-driven variation across languages. At the same time, it remains unclear whether this reflects a genuine robustness of universal syntax to language contact, or finer-grained syntactic phenomena beyond UPOS would reveal contact-related effects.

\begin{figure}
    \centering    
    \includegraphics[width=\linewidth]{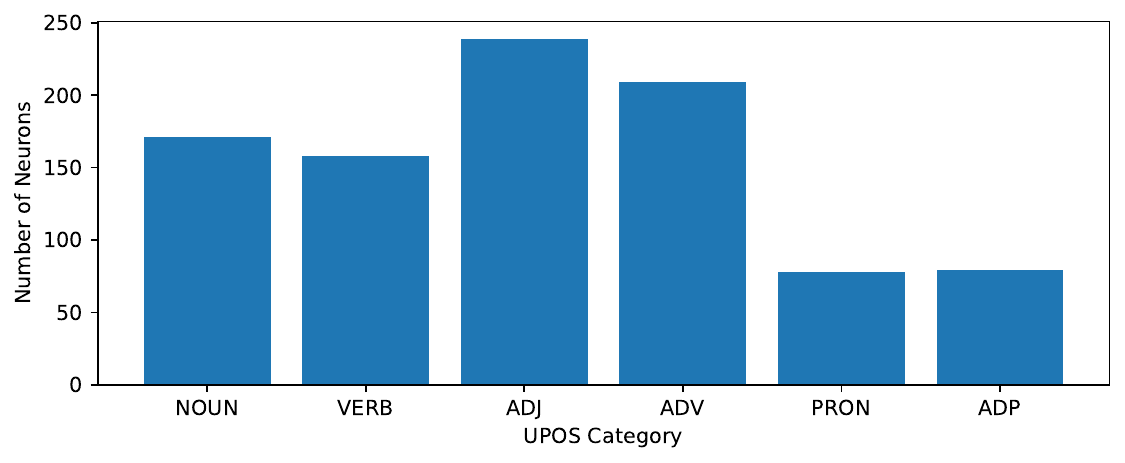}
    \caption{The LAPE scores for UPOS categories.}    
    \label{fig:upos_lape_total}
\end{figure}

Figure~\ref{fig:upos_lape_total} shows LAPE scores for each of the UPOS tags. The content-word categories (\textsc{ADJ}, \textsc{ADV}, \textsc{NOUN}, \textsc{VERB}) account for substantially more UPOS-selective neurons than the function-word categories \textsc{PRON} and \textsc{ADP}. Following \citet{tang-etal-2024-language}, larger LAPE scores indicate that the UPOS tag is captured by a larger set of \emph{highly selective} features. The large scores for content-word categories reflect their greater semantic variability and contextual dependence, which require more specialized features to be reliably distinguished. Function-word categories, however, encode abstract grammatical relations that are highly recurrent and structurally constrained, allowing them to be represented more compactly and with fewer isolated neurons. The patterns further reinforce the conclusion that universal syntactic categories are largely insensitive to language contact. 


\subsection{\textsc{Case}}
\begin{figure}
    \centering    \includegraphics[width=\linewidth]{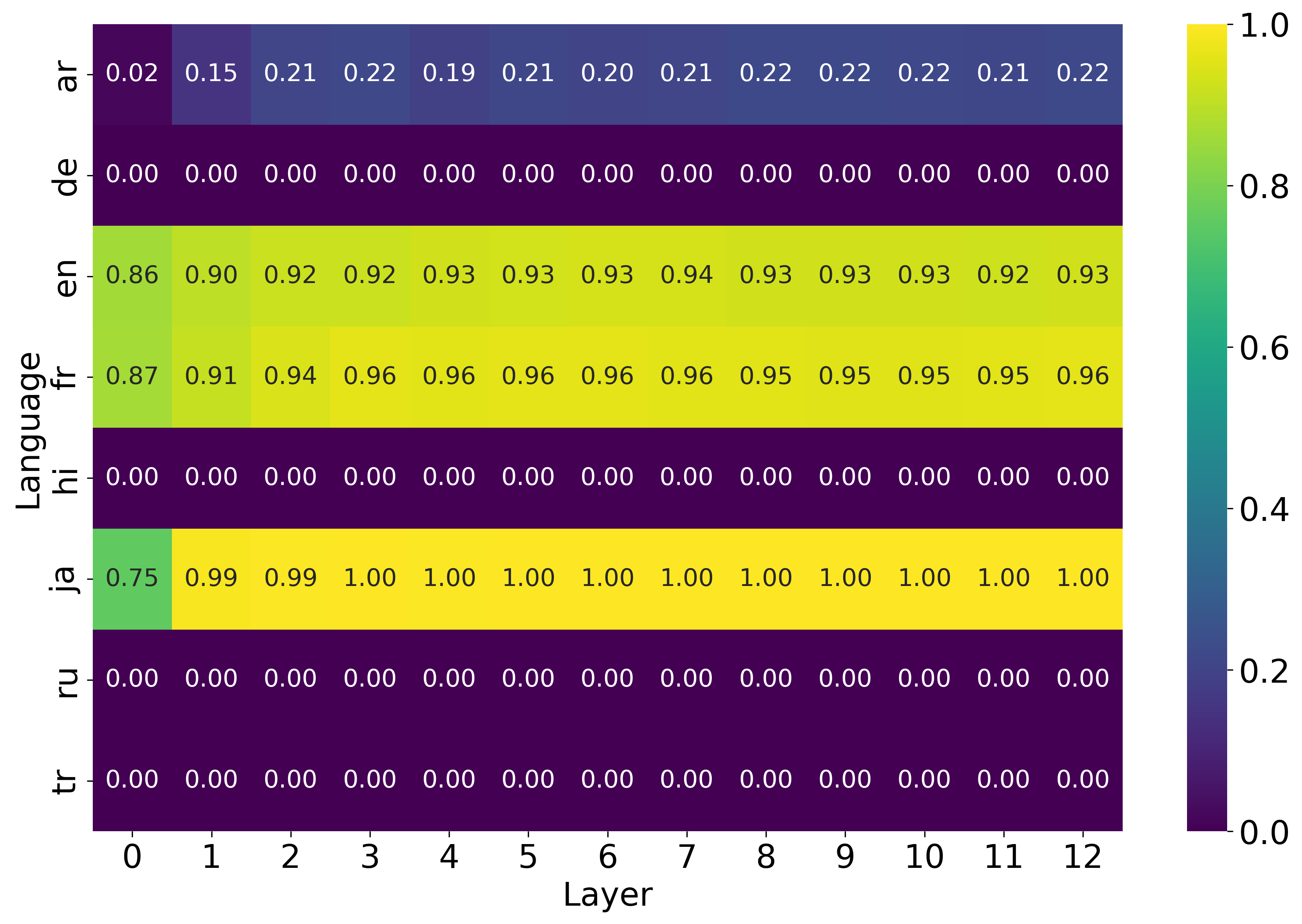}
    \caption{Normalized variational usable information ($\hat{I}_\mathcal{V}$) for \textsc{Case} prediction.}
    \label{fig:case_nuvi}
\end{figure}

Figure~\ref{fig:case_nuvi} reports the usable information values for the morphological feature \textsc{Case}. For languages with rich nominal case systems, namely German, Russian, and Turkish, the information gain remains close to zero throughout the layers, indicating that ParsBERT's representations provide no recoverable information about their case distinctions. In contrast, languages with limited or residual case marking, such as English, French, and Japanese, exhibit substantially higher usable information. For these languages, information gain increases sharply after the first layer and reaches values in the range of $0.9$--$1.0$ in the upper layers.

Arabic presents an intermediate pattern: although it possesses a morphologically rich case system, the usable information starts at approximately $0.02$ in the lowest layer and gradually increases to around $0.2$ in the final layers. 
This behavior is likely due to extensive lexical overlap between the two languages. Over centuries of lexical and semantic exchange, Persian may partially adopt patterns associated with Arabic-derived vocabulary, providing a modest boost alongside the structural cues it relies on. Moreover, the observation reflects the limited realization of case marking in contemporary written Arabic \citep{Fischer2002}, which partially aligns with modern Persian morpho-syntactic patterns. 

Among languages with limited case systems, Hindi exhibits consistently low information across all layers, suggesting that ParsBERT poorly captures its postpositional case. This likely stems from a structural mismatch between Hindi’s obligatory noun–postposition dependencies and Persian’s configurational encoding of argument structure, which relies on word order, prepositions, and discourse-level marking rather than local morphological cues. From a language-contact perspective, this reflects the predominantly lexical nature of Persian–Hindi contact, which has not extended to the transfer or convergence of core morphosyntactic patterns such as postpositional case systems.

Overall, the results show that ParsBERT encodes case-related information in a highly asymmetric manner, favoring languages with minimal or residual case marking while failing to capture rich inflectional case systems absent in Persian. This asymmetry can be due to the lack of inflectional case morphology in Persian, which encodes argument structure through rigid word order, differential object marking, adpositions, and dependency relations. Consequently, a Persian-trained model acquires abstract, structure-based representations of syntactic roles rather than paradigmatic case distinctions. When applied cross-lingually, this knowledge transfers more effectively to languages whose argument relations are encoded configurationally or analytically,\footnote{Here, \emph{configurational} refers to the encoding of grammatical relations through syntactic position and dependency structure (e.g., word order and argument–predicate configurations), rather than through overt morphological case marking.} such as English, French, and Japanese, and substantially less to languages in which case distinctions are realized through morphosyntactic marking, including Turkish, Russian, Hindi, German, and Arabic. This pattern underscores the sensitivity of morphological features to language-specific structure and historical development, in contrast to the relative robustness observed for universal syntactic categories.

\begin{figure}
    \centering    \includegraphics[width=\linewidth]{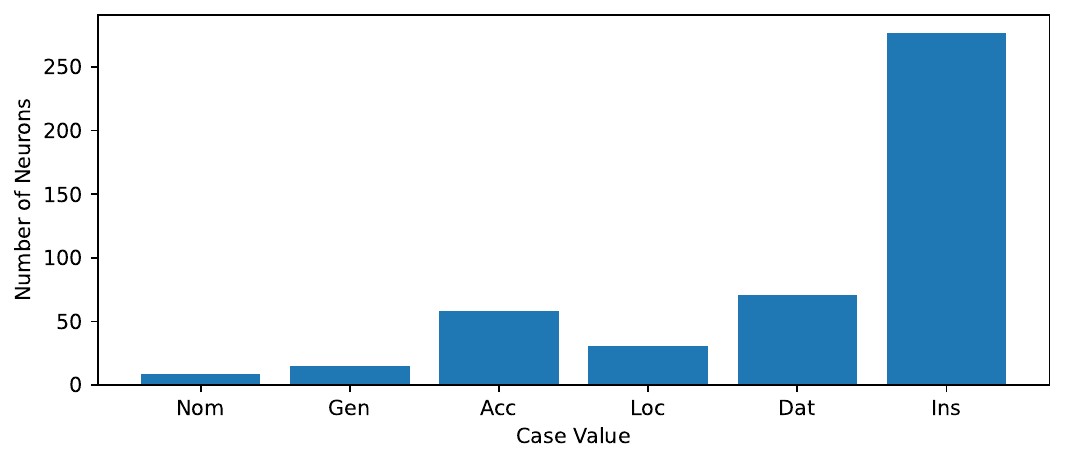}
    \caption{The LAPE scores for \textsc{Case} categories.}    \label{fig:case_lape_total}
\end{figure}

Figure~\ref{fig:case_lape_total} reports the LAPE scores for individual \textsc{Case} categories. The prominence of \textsc{Ins} and \textsc{Dat} indicates that ParsBERT lacks compact, category-specific features for these relations and instead relies on a distributed set of weak structural cues. This aligns with the fact that instrumental and dative relations in Persian are typically expressed analytically through adpositions, light-verb constructions, and stable dependency patterns rather than via inflectional morphology. Consequently, when these relations are realized morphologically in other languages, the model must reconstruct them indirectly, resulting in higher LAPE scores and reduced representational efficiency.

In contrast, the comparatively low scores for \textsc{Nom} and \textsc{Acc} reflect their close structural alignment with Persian’s encoding of core argument structure. Subject and object roles are strongly constrained by syntactic position in Persian, allowing these relations to be represented compactly with fewer selective features, even in the absence of overt case marking. From a language-contact perspective, these results suggest that while contact has reinforced analytic strategies for expressing non-core grammatical relations, it has not led to the transfer of inflectional case paradigms, shaping both the availability and efficiency of case representations in the model.

\subsection{\textsc{Gender}}
\begin{figure}
  \centering
    \includegraphics[width=\linewidth]{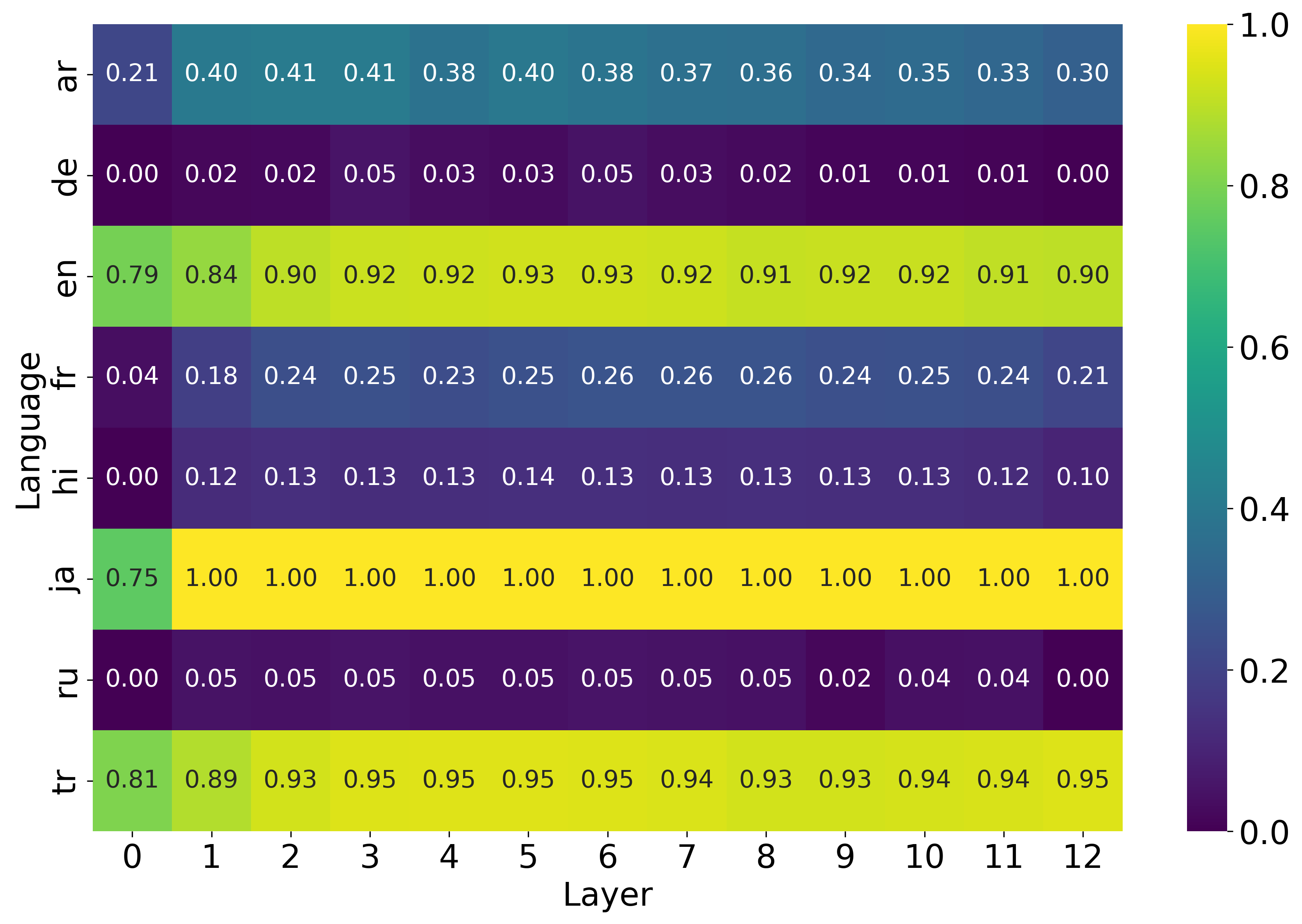}
    \caption{Normalized variational usable information for \textsc{Gender} prediction.}
    \label{fig:gender_nuvi}
\end{figure}

Figure~\ref{fig:gender_nuvi} represents the values of usable information for \textsc{Gender} across languages and layers. High usable-information for gender-neutral languages such as Japanese and Turkish, as well as for English, whose gender marking is largely restricted to pronominal forms, indicate that ParsBERT captures cues associated with the absence or marginal realization of grammatical gender, consistent with Persian’s gender-neutral system. However, languages with fully grammaticalized gender systems exhibit lower and variable results. No recoverable information is available for three-gender languages, whereas two-gender languages, such as French, Hindi, and Arabic, exhibit moderate usable information, with Arabic resulting in the highest usable information among gendered languages, likely reflecting partial surface-level alignment and extensive lexical overlap with Persian. 

\begin{figure}
    \centering    \includegraphics[width=\linewidth]{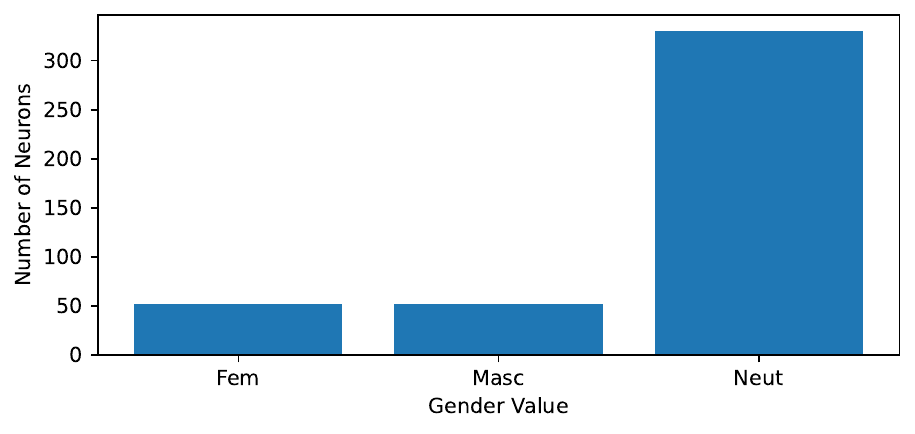}
    \caption{The LAPE scores for Gender categories.}    \label{fig:gender_lape_total}
\end{figure}

Figure~\ref{fig:gender_lape_total} reports LAPE scores for each gender category. Gender-related selectivity in ParsBERT is concentrated in \textsc{Neut}-associated features. This observation is consistent with the usable-information results for three-gender systems and likely reflects the greater semantic diversity and lexical heterogeneity of neuter nouns. Because neuter nouns typically encompass a broad range of inanimate and abstract referents, a Persian-trained model distributes their representation across many features, each capturing a subset of the neuter class, whereas masculine and feminine nouns, which are often more semantically constrained, can be encoded with fewer selective features. 

Overall, the better performance of two-gendered languages, compared to three-gendered ones, can be explained from two perspectives: the nature of the gender systems themselves and the type of language contact Persian has experienced. From a structural standpoint, two-gender systems tend to align more closely with general semantic and discourse cues such as animacy, humanness, and natural sex, which remain accessible even in a gender-neutral language like Persian and can therefore be exploited by ParsBERT. In contrast, three-gender systems rely on largely arbitrary noun-class assignments and agreement paradigms that must be learned morphologically and are not recoverable from syntactic configuration or semantics alone. 

From a language-contact perspective, this asymmetry is further shaped by the predominantly lexical and semantic nature of Persian’s contact with gendered languages, which facilitates surface-level alignment and the recoverability of semantically grounded gender distinctions typical of two-gender systems, without inducing the grammatical restructuring necessary to support fully paradigmatic gender systems.

\section{Conclusion}

We investigated whether contact-induced linguistic structure is reflected in the internal representations of a monolingual Persian-trained language model. The results reveal a clear asymmetry between syntactic and morphological features: universal syntactic categories (UPOS) show little sensitivity to historical contact, whereas morphological features such as \textsc{Case} and \textsc{Gender} are strongly shaped by language-specific structure. In particular, the model favors analytic and configurational encodings aligned with Persian, while failing to capture rich inflectional paradigms absent from the training language. The attribution analysis further indicates that contact-related alignment is primarily reflected through distributed cues rather than compact, category-specific features, pointing to surface-level or semantic alignment rather than grammatical transfer. Overall, these findings suggest that monolingual language models implicitly encode selective traces of language contact, constrained by the structural properties of the training language. 

\section*{Limitations}
A central limitation of our study is that the PUD collection does not include several languages that have played crucial roles in the historical and areal development of Persian. Notably absent are major Turkic varieties such as Azerbaijani and Turkmen, which have exerted long-term and deeply structural influence on Persian, often more directly than modern Turkish. Similarly, Urdu, the closest Indo-Aryan successor to the Persianate linguistic tradition and a primary locus of Persian lexical and syntactic transfer, is missing from PUD, preventing a fuller assessment of Persian's influence in South Asia. Important regional contact languages such as Armenian and Kurdish, both of which share extensive areal features with Persian, are also unavailable. The exclusion of these languages constrains the breadth of our analysis, as the strongest cases of sustained bilingualism, bidirectional borrowing, and structural convergence cannot be directly evaluated within the PUD framework. Future work incorporating UD treebanks or comparable resources for these languages would enable a more comprehensive and historically aligned investigation of Persian language contact.


\bibliography{custom}

\appendix

\section{Layer-wise Distribution of LAPE}
\label{appx:lape}
In this section, we provide an overview of the LAPE scores across layers and task categories. Detailed analyses for each task are presented in the following subsections. 

\subsection{Language Identification}

\begin{figure}
    \centering
    \includegraphics[width=\linewidth]{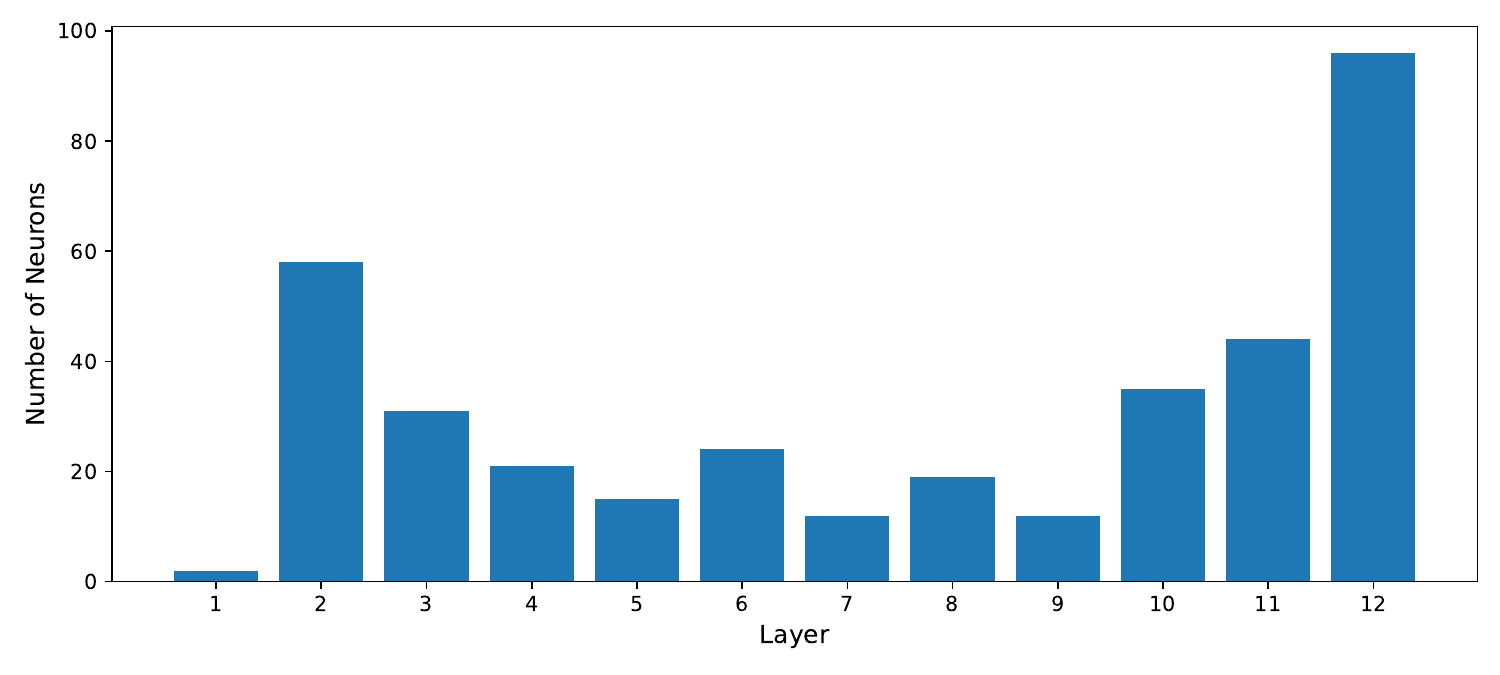}
    \caption{Layer-wise distribution of language-specific neurons identified by LAPE.}
    \label{fig:lang_id_lape}
\end{figure}

Figure~\ref{fig:lang_id_lape} shows the total number of language-specific neurons identified at each layer. 
The results show that language-specific neurons are concentrated in a small number of layers, most notably Layer~2 and the top layers, rather than being evenly distributed throughout the network.

\begin{figure}
    \centering
    \includegraphics[width=\linewidth]{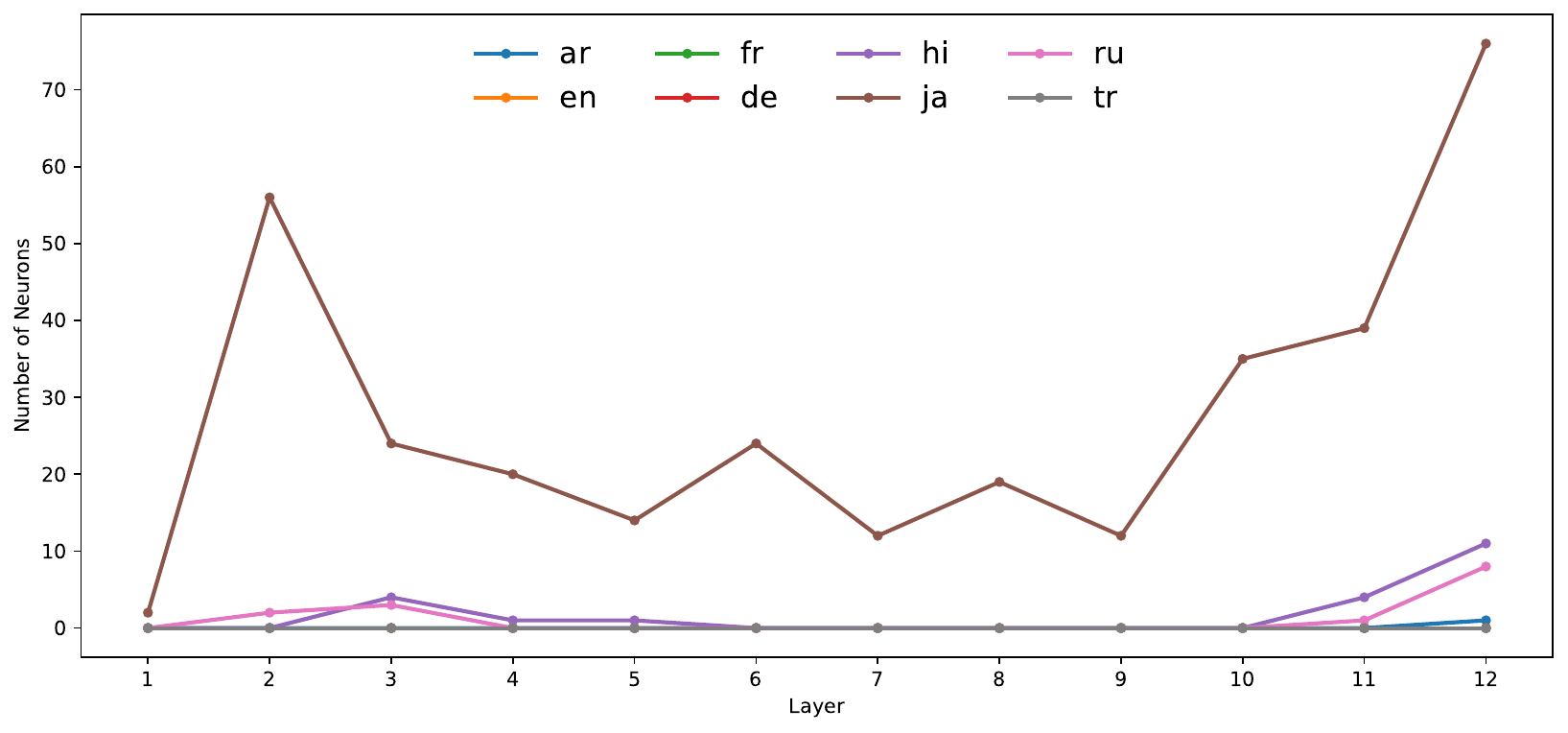}
    \caption{Layer-wise distribution of language-specific neurons identified by LAPE.}
    \label{fig:lang_id_lape_line}
\end{figure}

Figure~\ref{fig:lang_id_lape_line} further breaks down the results by language. The dominance of Japanese-specific neurons across almost all layers, particularly in the upper layers, indicates that ParsBERT allocates a large number of highly specialized neurons to processing Japanese tokens. This likely reflects strong script-level and orthographic differences between Japanese and Persian, which require distinct representational pathways despite the model’s monolingual training. 

Hindi and Russian exhibit a more moderate but still noticeable concentration of language-specific neurons, primarily in higher layers. This aligns with the results from usable information, where both languages show relatively high recoverable language identity information. 
In contrast, English, French, German, and Turkish are associated with very few or no language-specific neurons. This suggests that for these languages, language identity is either weakly encoded or distributed across shared neurons rather than being localized in highly specific units.

\subsection{UPOS Identification}

\begin{figure}
    \centering    \includegraphics[width=\linewidth]{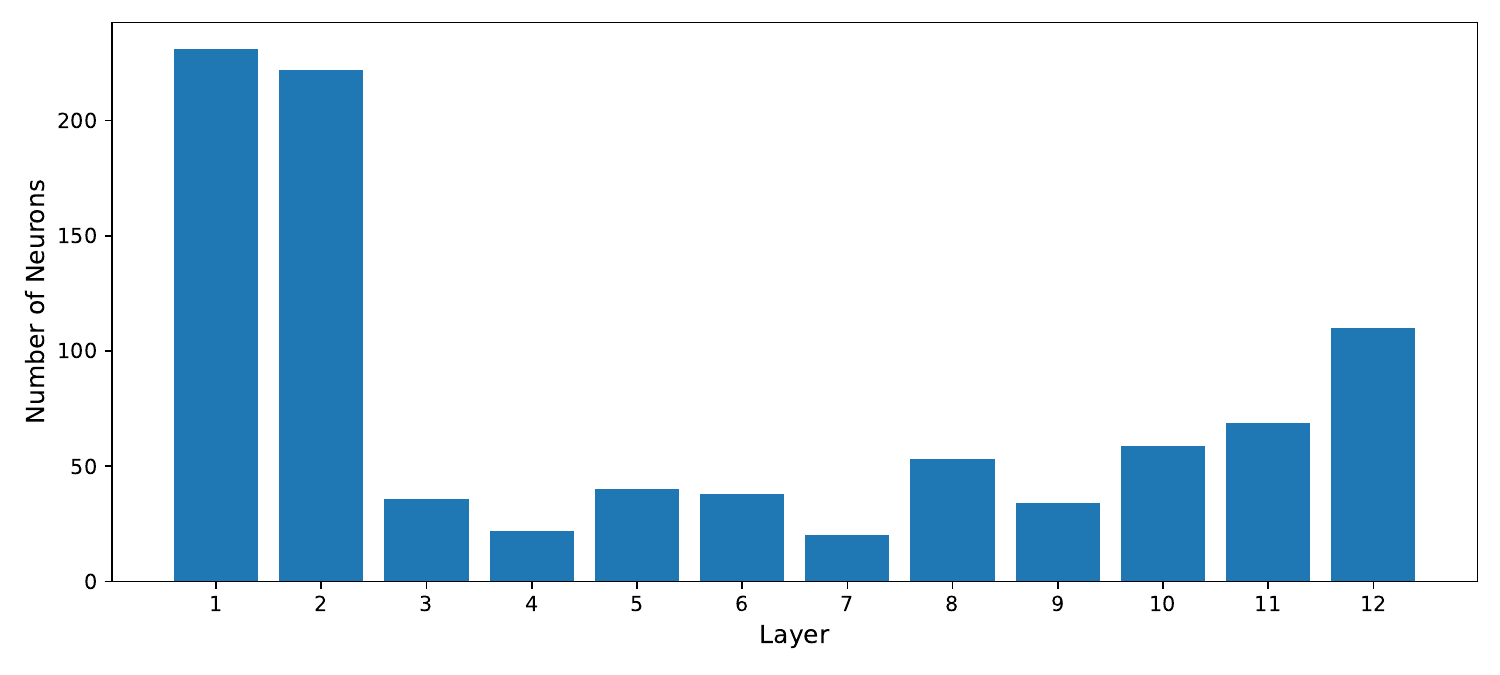}
    \caption{Layer-wise distribution of UPOS-specific neurons identified by LAPE.}
    \label{fig:upos_lape}
\end{figure}

Figure~\ref{fig:upos_lape} shows the layer-wise LAPE scores for UPOS identification. 
The LAPE results indicate that UPOS selectivity is present both very early and again in the upper part of the network, with weaker neuron-level selectivity in the intermediate layers.

\begin{figure}
    \centering    
    \includegraphics[width=\linewidth]{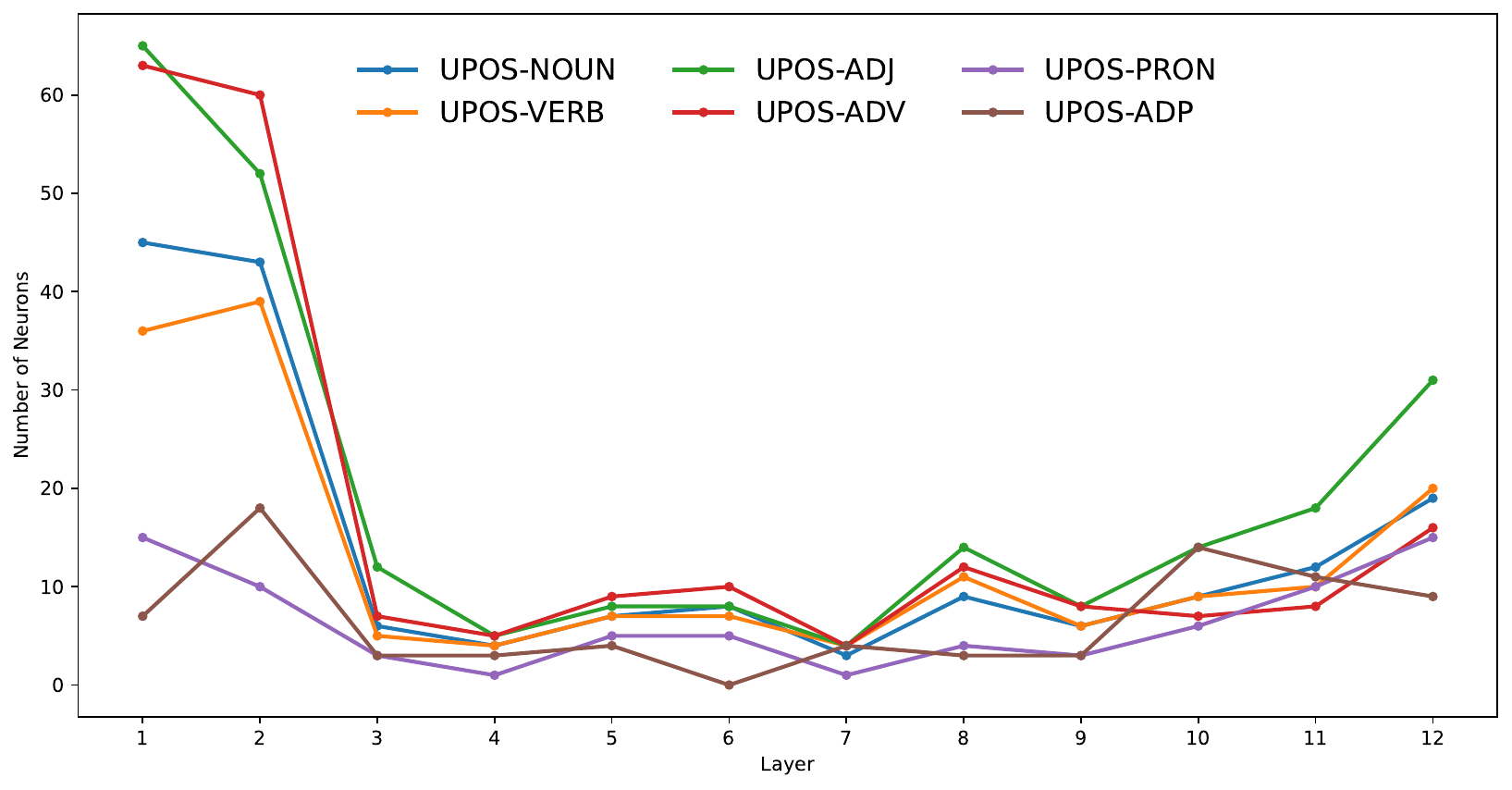}
    \caption{Layer-wise distribution of UPOS-specific neurons identified by LAPE.}
    \label{fig:upos_lape_line}
\end{figure}

The per-category breakdown in Figure~\ref{fig:upos_lape_line} shows that \textsc{ADJ} and \textsc{ADV} contribute the largest numbers of UPOS-selective neurons across layers, followed by \textsc{NOUN} and \textsc{VERB}. In contrast, \textsc{PRON} and \textsc{ADP} contribute substantially fewer selective neurons throughout the network. This indicates that the overall LAPE profile is driven primarily by content-word categories, whereas function-word categories yield comparatively fewer neurons that meet the LAPE selectivity criterion. 
From the perspective of language contact, this asymmetry may indicate that ParsBERT encodes function-word categories in a more abstract and transferable manner across languages, whereas content-word representations are more language-specific and therefore less directly aligned across Persian and the test languages.

\subsection{Case}

\begin{figure}
    \centering
    \includegraphics[width=\linewidth]{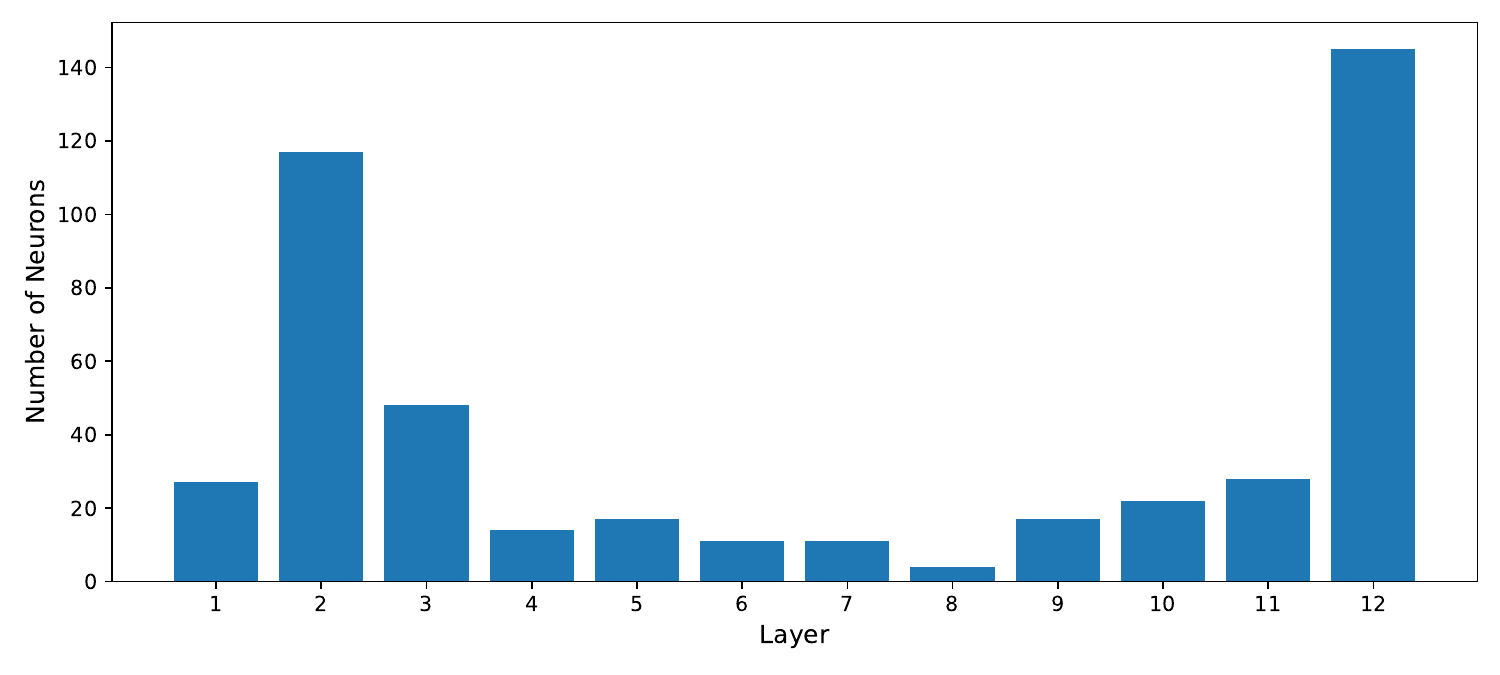}
    \caption{Layer-wise distribution of Case-specific neurons identified by LAPE.}
    \label{fig:case_lape}
\end{figure}

\begin{figure}
    \centering    \includegraphics[width=\linewidth]{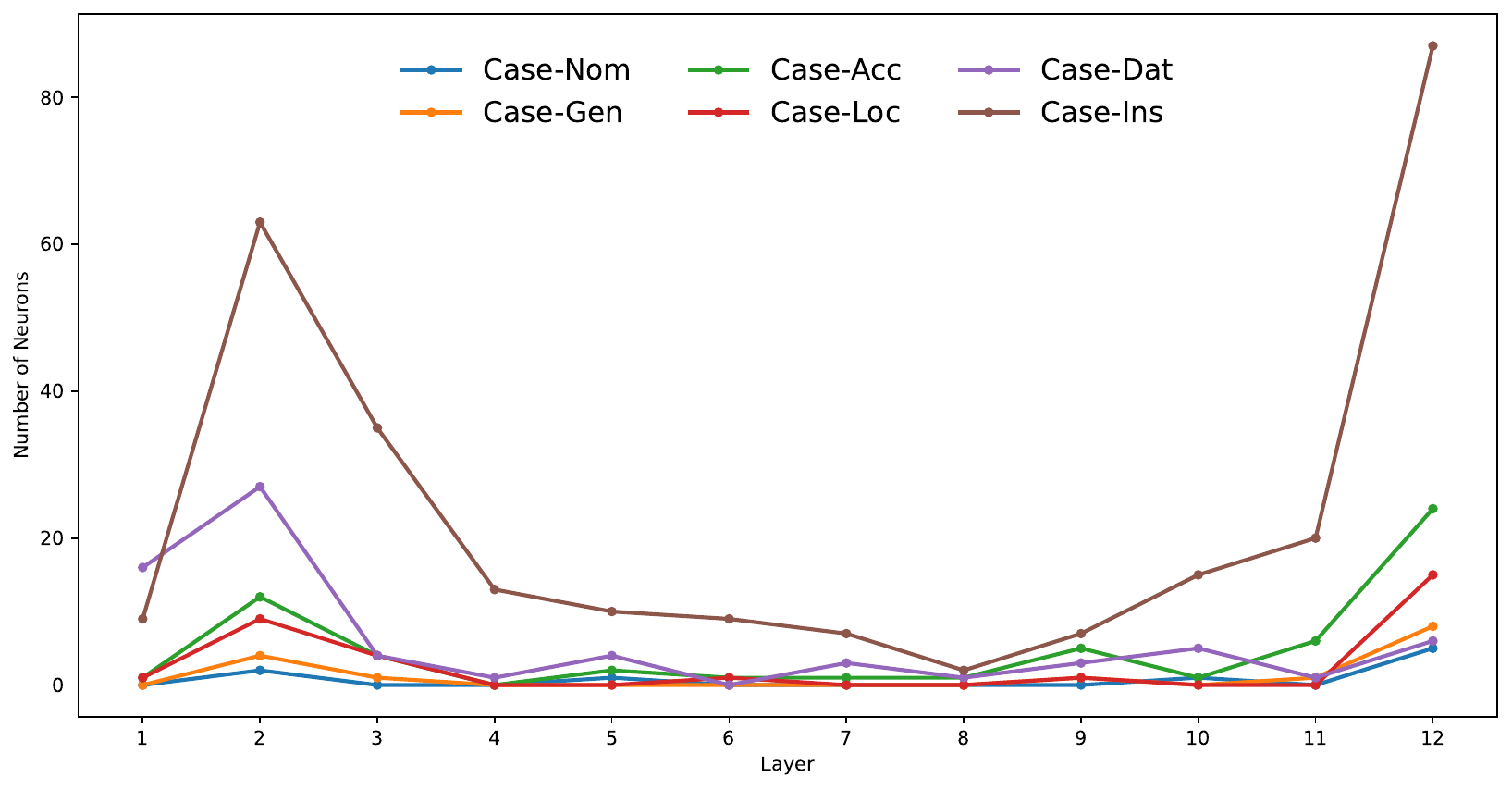}
    \caption{Layer-wise distribution of Case-specific neurons identified by LAPE.}
    \label{fig:case_lape_line}
\end{figure}

Figure~\ref{fig:case_lape} presents the layer-wise distribution of Case-specific neurons, and Figure~\ref{fig:case_lape_line} further decomposes this distribution by individual case values. The number of selected neurons is low in Layer~1, peaks sharply in Layer~2, drops substantially across the middle layers, and then increases again toward the final layer, where it reaches its maximum.
This indicates that case selectivity is concentrated in a small subset of layers rather than being evenly distributed across the network depth.

The per-case breakdown in Figure~\ref{fig:case_lape_line} shows that this pattern is driven primarily by \textsc{Ins}, which accounts for the largest share of case-selective neurons and exhibits pronounced peaks in Layers~2 and~12.
Other case values contribute far fewer neurons overall: \textsc{Dat} shows a moderate peak in the early layers, while \textsc{Acc} and \textsc{Loc} increase mainly in the final layer; \textsc{Nom} and \textsc{Gen} remain sparse throughout.
Overall, the LAPE analysis indicates that, when case-related selectivity is observed, it is localized to a limited number of layers and concentrated on a small subset of case values.

\subsection{Gender}

\begin{figure}
    \centering
    \includegraphics[width=\linewidth]{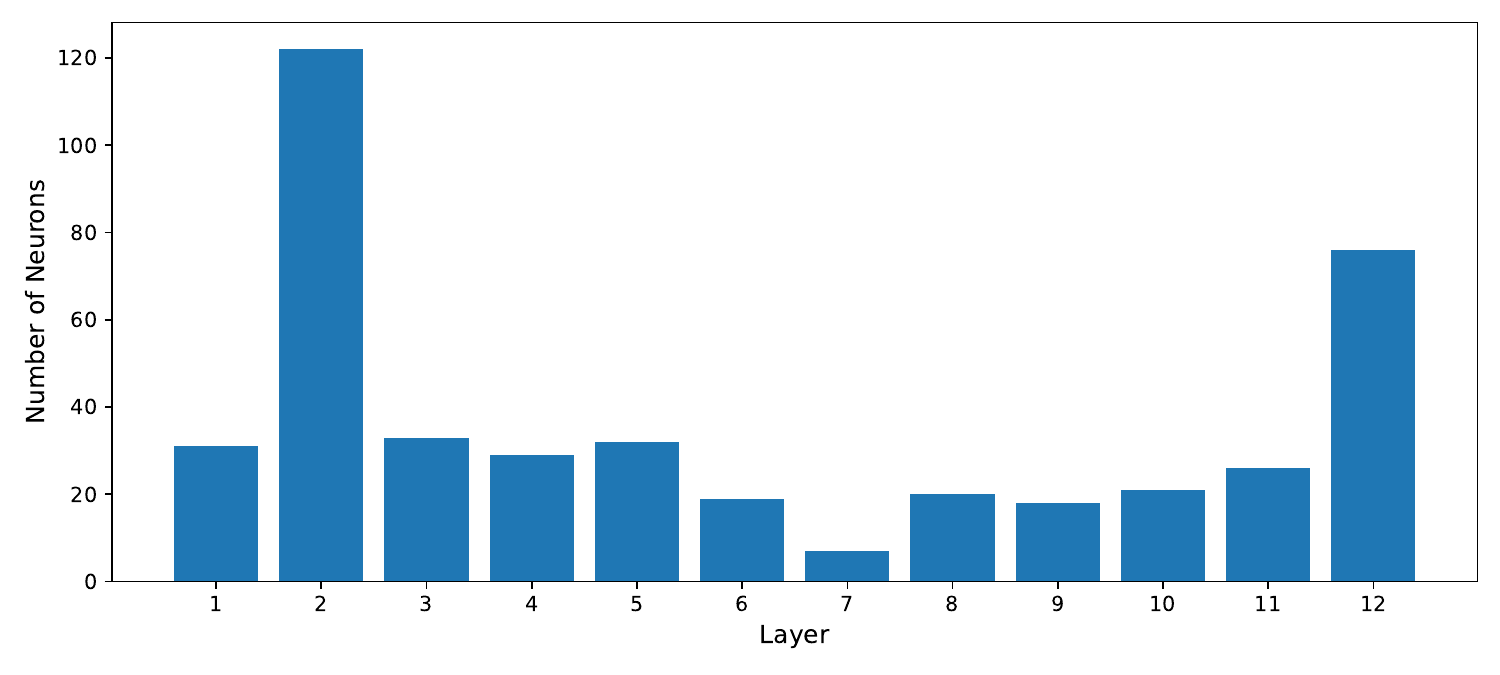}
    \caption{Layer-wise distribution of Gender-specific neurons identified by LAPE.}
    \label{fig:gender_lape}
\end{figure}

Figure~\ref{fig:gender_lape} shows the layer-wise distribution of gender-selective neurons identified by LAPE.
The results show that gender selectivity is not uniformly distributed across depth, but concentrates in a small number of layers, especially early (Layer~2) and late (Layer~12).

\begin{figure}
    \centering    \includegraphics[width=\linewidth]{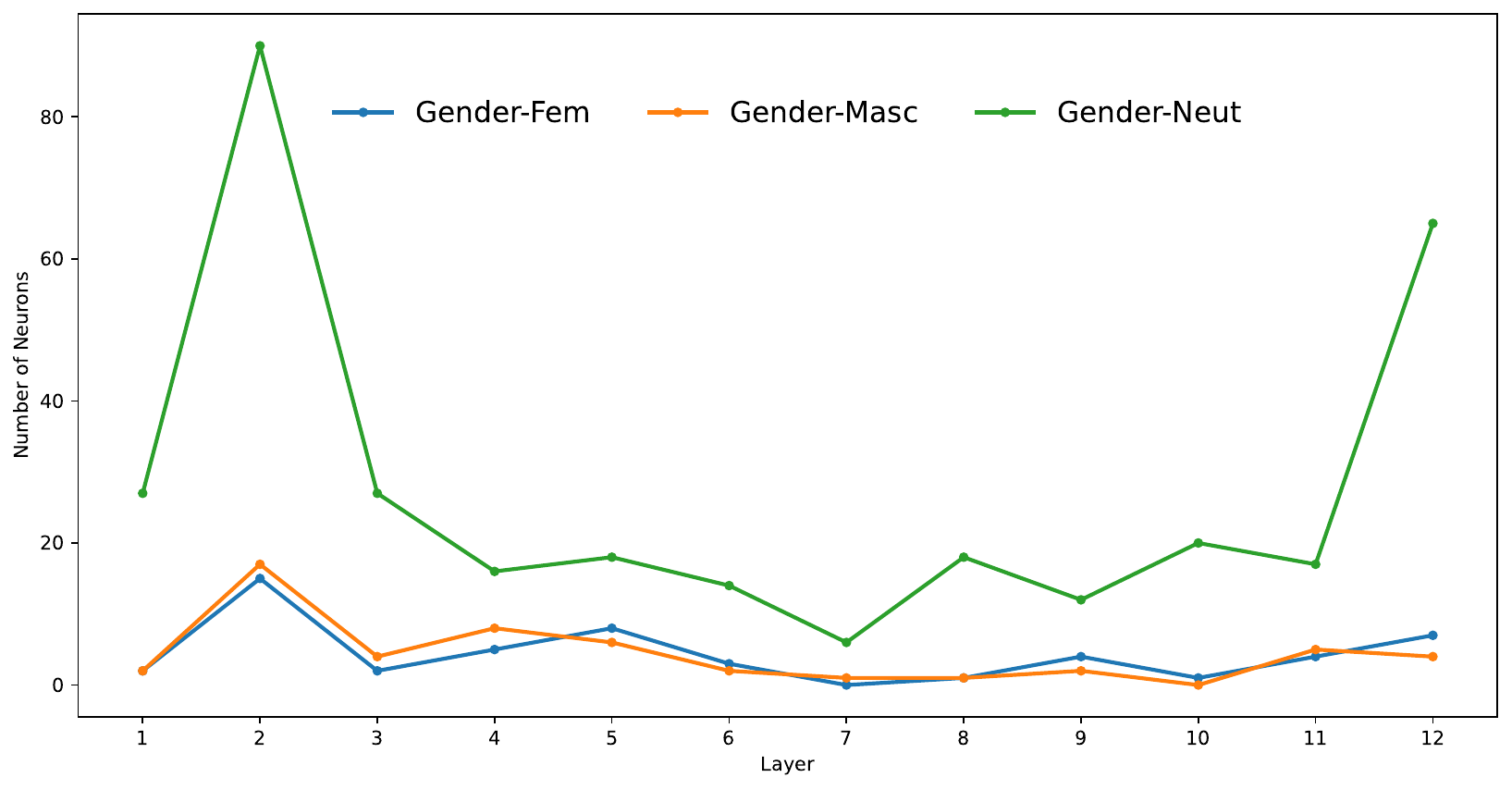}
    \caption{Layer-wise distribution of Gender-specific neurons identified by LAPE.}
    \label{fig:gender_lape_line}
\end{figure}

The per-category breakdown in Figure~\ref{fig:gender_lape_line} indicates that the aggregate pattern is driven primarily by the \textsc{Neut} category, which accounts for the majority of gender-selective neurons in nearly every layer and shows strong peaks in Layers~2 and~12.
In contrast, \textsc{Fem} and \textsc{Masc} contribute comparatively few neurons and remain low across the network, with only small increases in the upper layers.
Overall, the LAPE analysis suggests that neuron-level selectivity for gender is dominated by the \textsc{Neut} label, while selectivity for \textsc{Fem}/\textsc{Masc} distinctions is comparatively sparse.

\end{document}